\documentclass[runningheads]{llncs}
\usepackage[T1]{fontenc}
\usepackage{graphicx}
\usepackage{amsmath} 
\usepackage{amssymb}  
\usepackage{tikz,booktabs}
\usepackage{pgf}
\usepackage{tzplot}
\usepackage{pgfplots}
\usepackage{float}
\usepackage[colorlinks=true, allcolors=blue]{hyperref}
\usepackage[capitalise]{cleveref}
\usepackage{wrapfig}
\usepackage[caption=false]{subfig}
\usepackage{paralist}
\usepackage{floatflt}

\usepackage{environ}
\makeatletter
\newsavebox{\measure@tikzpicture}
\NewEnviron{scaletikzpicturetowidth}[1]{%
  \def\tikz@width{#1}%
  \begin{lrbox}{\measure@tikzpicture}%
  \BODY
  \end{lrbox}%
  \pgfmathparse{#1/\wd\measure@tikzpicture}%
  \BODY
}
\makeatother

\definecolor{darkgreen}{RGB}{0,100,0}

\begin{document}
%
%
\title{Split over $n$ resource sharing problem: Are fewer capable agents better than many simpler ones?}

%
%
\titlerunning{The split over $n$ resource sharing problem}
%
\author{Karthik Soma\inst{1,2}\orcidID{0009-0009-8411-6679} \and
Mohamed S. Talamali\inst{2}\orcidID{0000-0002-2071-4030} \and
Genki Miyauchi\inst{2}\orcidID{0000-0002-3349-6765}  \and
Giovanni Beltrame\inst{1}\orcidID{0000-0001-9755-8630}  \and
Heiko Hamann\inst{3}\orcidID{0000-0002-2458-8289}
\and
Roderich Gro{\ss}\inst{2,4}\orcidID{0000-0003-1826-1375}
}


%
\authorrunning{K. Soma et al.}
%
\institute{
MIST lab, École Polytechnique de Montréal, Montreal, Canada\\ 
\email{karthik.soma@polymtl.ca}
\and 
School of Electrical and Electronic Engineering, The University of Sheffield, Sheffield, UK\\
\and 
Department of Computer and Information Science, University of Konstanz, Konstanz, Germany 
\and
Department of Computer Science, Technical University of Darmstadt, Darmstadt, Germany\\
}
%
\index{Soma, Karthik}
\index{Talamali, Mohamed S.}
\index{Miyauchi, Genki}
\index{Beltrame, Giovanni}
\index{Hamann, Heiko}
\index{Gro{\ss}, Roderich}
%
\maketitle              
\begin{abstract}

In multi-agent systems, should limited resources be concentrated into a few capable agents or distributed among many simpler ones? This work formulates the split over $n$ resource sharing problem where a group of $n$ agents equally shares a common resource (e.g., monetary budget, computational resources, physical size). We present a case study in multi-agent coverage where the area of the disk-shaped footprint of agents scales as $1/n$. A formal analysis reveals that the initial coverage rate grows with $n$. However, if the speed of agents decreases proportionally with their radii, groups of all sizes perform equally well, whereas if it decreases proportionally with their footprints, a single agent performs best. We also present computer simulations in which resource splitting increases the failure rates of individual agents.
The models and findings help identify optimal distributiveness levels and inform the design of multi-agent systems under resource constraints.
\end{abstract}

\section{Introduction}


Artificial systems must operate with limited resources. These resources could be concentrated into a single agent. Alternatively, they could be divided among multiple agents. For example, consider a system designer with a limited monetary budget to purchase robots for a given mission. If opting for a single-agent system, they could afford a high-spec robot featuring advanced actuation, sensing, and computational resources. However, if opting for a system comprising many agents, they could only afford robots of substantially lower cost (per unit) featuring lower quality actuation, sensing, and computational resources. 

\begin{figure}
    \centering
    \includegraphics[width=0.6\linewidth]{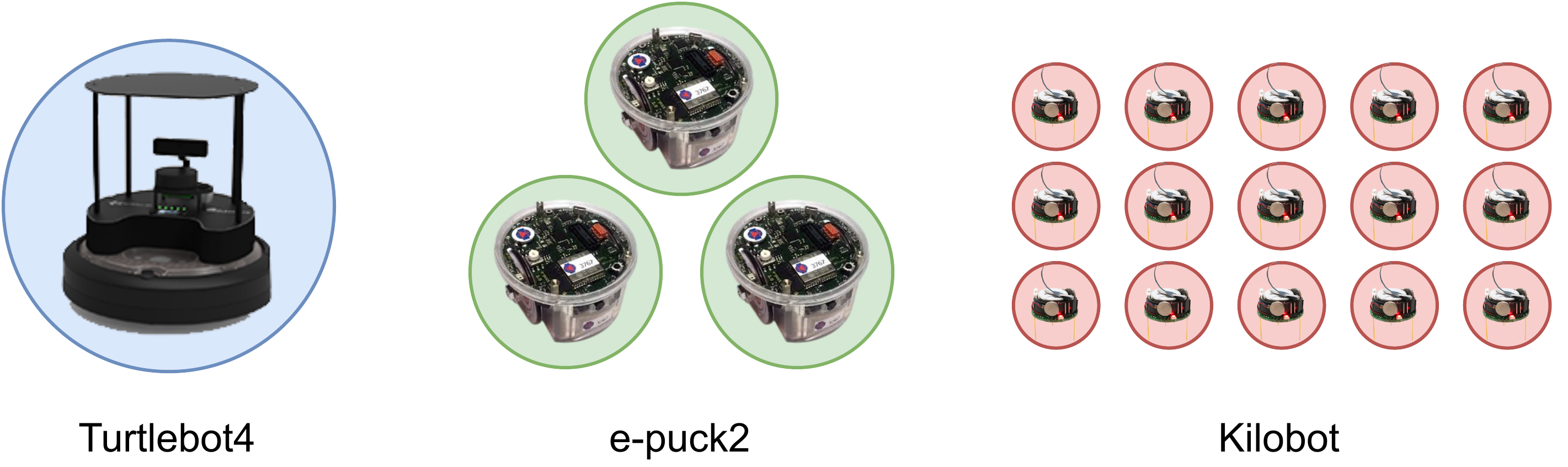}
    \caption{Split over $n$ resource sharing problem. A finite resource is equally shared among $n$ robots; What is the optimal $n$? Different values of $n$ might correspond to different robot platforms, for example, Turtlebot4~\cite{turtlebot20}, e-puck2~\cite{epuck09}, and Kilobot~\cite{Kilobot12}.}
    \label{fig:problem-overview}
\end{figure}
This paper formulates the aforementioned dilemma as the \emph{split over $n$ resource sharing problem} (see Fig.~\ref{fig:problem-overview}), where $n$ agents are assumed to each possess an equal share of a limited resource. The problem addresses a fundamental question in swarm intelligence: given a fixed resource, what is the optimal level of distributiveness? 


The split over $n$ resource sharing problem differs from the vast literature in the field of swarm intelligence that focuses on the optimal group size under the assumption of an effectively unlimited supply of identical robots. For example, will 5, 50, or 500 robots of a particular design yield the best performance? Such  studies~\cite{HeikoSuper18,HamannScalability22} provide models and answers for questions about optimal group size or swarm density (robots/area). These models show an initial increase of system performance with increasing number of robots and robot density that can even be superlinear, that is, doubling the number of robots results in more than doubled system performance~\cite{HeikoSuper18}.
There is an optimal number of robots (and density) at which system performance is maximized. Increasing the number of robots beyond this critical value leads to performance degradation due to congestion or interference~\cite{SomaCongestion23}. 
This paper investigates the split over $n$ resource sharing problem in the context of a spatial coverage task \cite{Choset2001:AnnalsMathAI,Alcherio-coverage-DARS,10665968}, where agents must visit all locations of the environment collectively. Spatial coverage tasks are fundamental to many real-world applications, such as vacuuming floors and mapping.
As coverage depends on both footprint and mobility, it serves as a benchmark for this study.

The contributions of this work are threefold. First, we formulate the split over $n$ resource sharing problem, the solution to which may inform optimal level of distributiveness in multi-robot system design assuming a constant resource budget. Second, we present a multi-agent coverage case study incorporating practical miniaturization constraints, such as reduced mobility and higher failure rates. Third, we formally model the system and reveal conditions under which the optimal level of distributiveness is minimal, maximal, or irrelevant.

\section{Formulation of the Split Over $n$ Resource Sharing Problem}~\label{sec:problem}
We assume a periodic environment of size $p \times p$ arbitrary units. It is populated by $n$ identical, disk-shaped agents. Let $A>0$ denote the shared resource, representing the combined footprint of all agents. Hence, the resource (i.e., footprint) per agent is $A/n$. 
%
%
Each agent has a continuous position in 2-D space and an orientation. It propels by alternating between (i)~moving forward with velocity $v$ for $\delta$ time steps, and (ii)~rotating instantaneously by
angle $\theta$ (in rad).
$\delta$ is drawn from a power-law distribution and turning angles $\theta$ from a wrapped Cauchy distribution to realize a random walk~\cite{DimidovRandomWalk16}, given by:

\noindent
\begin{minipage}{0.4\textwidth}
\begin{equation*}
    P_{\alpha}(\delta) \ \propto \delta^{-(\alpha + 1)}, \ 0< \ \alpha \ \leq 2 \;.
\end{equation*}
\end{minipage}
\hfill
\begin{minipage}{0.5\textwidth}
\begin{equation*}
    f_w(\theta;\mu,\rho) = \frac{1}{2\pi}\frac{1-\rho^2}{1+\rho^2-2\rho\cos{(\theta-\mu)}}\;\;
\end{equation*}
\end{minipage}

We consider Brownian random motion~\cite{DimidovRandomWalk16} ($\rho=0,\ \mu=0,\ \alpha=2$) and investigate four velocity profiles that capture practical conditions where footprint miniaturization imposes limits on the maximum achievable velocity of smaller sized agents (see~\cref{fig:vel_prof}):

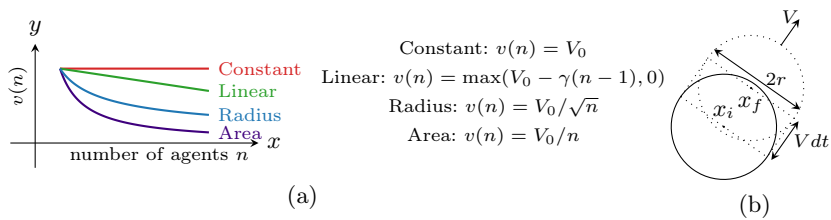
\begin{figure}[tbp]
  \centering
  \subfloat[][]{
    \begin{minipage}{0.75\textwidth}
    \noindent
    \centering
    \begin{tabular}{@{}cc}
      \begin{minipage}{0.45\textwidth}
        \centering
        \definecolor{mtlred}{HTML}{D62728}
        \definecolor{mtlpurple}{HTML}{3F007D}
        \definecolor{mtlgreen}{HTML}{2CA02C}
        \definecolor{mtlblue}{HTML}{1F77B4}
        \begin{tikzpicture}[scale=0.33]
          \tzaxes(-1,-1)(9,4){$x$}{$y$}
          \node at (5,-0.5) {\scriptsize number of agents $n$};
          \node[rotate=90] at (-0.7,2.3) {\scriptsize $v(n)$};
          \tzfn[mtlred,thick]{3}[1:7]{\scriptsize Constant}[r]
          \tzfn[mtlgreen,thick]{3 - 0.15*(\x-1)}[1:7]{\scriptsize Linear}[r]
          \tzfn[mtlpurple,thick]{3*(\x)^-1}[1:7]{\scriptsize Area}[r]
          \tzfn[mtlblue,thick]{3*(\x)^-0.5}[1:7]{\scriptsize Radius}[r]
        \end{tikzpicture}
      \end{minipage}
      &
      \begin{minipage}{0.4\textwidth}
        \centering
        \begin{tabular}{c}
          \scriptsize Constant: $v(n) = V_0$ \\ 
          \scriptsize Linear: $v(n) = \max(V_0 - \gamma(n-1),0)$ \\
          \scriptsize Radius: $v(n) = V_0/\sqrt{n}$ \\
          \scriptsize Area: $v(n) = V_0/n$ 
        \end{tabular}
      \end{minipage}
    \end{tabular}
    \label{fig:vel_prof}
  \end{minipage}
  }
  \subfloat[][]{
    \begin{minipage}{0.2\textwidth}
    \centering
    \begin{tikzpicture}[scale=0.7,rotate=55]
      \draw[color=black] (0,0) circle [radius=1];
      \draw[black,dotted] (0.9,0) circle (1);
      \draw[black,dotted] (0,1) rectangle (0.9,-1);
      \draw[->] (1.9,0) -- (2.5,0) node[midway,above,anchor=south,align=center] {\scriptsize $V$};
      \draw[<->] (0,-1.1) -- (0.9,-1.1) node[midway,right] {\scriptsize $Vdt$};
      \draw[<->] (1.1,1) -- (1.1,-1) node[midway,right] {\scriptsize $2r$};
      \coordinate [label=90:$x_{i}$ ]  ($x_{i}$) at (0,0) ;
      \coordinate [label=-90:$x_{f}$ ]  ($x_{f}$) at (0.9,0) ;
      \draw[color=black] (0,0) circle [radius=0.01];
      \draw[color=black] (0.9,0) circle [radius=0.01];
    \end{tikzpicture}
    \label{fig:init-coverage}
  \end{minipage}
  }
  \caption{(a) The four velocity profiles as functions of $n$. (b) Area covered by a disk-shaped agent of radius $r$ moving forward with velocity $V$.}
  \label{fig:main}
\end{figure}





\begin{compactenum}
    \item \textit{constant}: an agent's velocity is constant, $V_0$, that is, irrespective of the group size~($n$);
    \item \textit{linear}: an agent's velocity is $V_0$ if $n=1$, however, as $n$ increases, it decreases proportionally until reaching zero;
    \item \textit{radius}: an agent's velocity is proportional to $\frac{1}{\sqrt{n}}$. Hence, it is proportional to the radius of its footprint, which is $\sqrt{\frac{A}{n\pi}}$;
    \item \textit{area}: an agent's velocity is proportional to $\frac{1}{n}$. Hence, it is proportional to the area of its footprint, which is $\pi \sqrt{\frac{A}{n\pi}} \sqrt{\frac{A}{n\pi}} = \pi \frac{A}{n\pi}=\frac{A}{n}$.
\end{compactenum}

To determine the area covered, we discretize the environment into $m \times m$ square cells. At any given time, an agent senses all cells whose centers lie within its footprint.
%

The coverage percentage up to time $t$ is given as:

\noindent
\begin{minipage}{0.48\textwidth}
\begin{equation*}
 c(t)=\frac{\sum_{i=1}^{m} \sum_{j=1}^{m} P_{i,j}(t)}{m^2}\cdot100
\end{equation*}
\end{minipage}
\hfill
\begin{minipage}{0.48\textwidth}

\begin{equation*}
 P_{i,j}(t) = 
\begin{cases}
    1 & \text{if cell } (i,j) \text{ covered}\\
    0 & \text{otherwise}
\end{cases}
\end{equation*}
\end{minipage}

\noindent
Let $t_f$ denote the first time at which the coverage reaches $100\%$, where $t_f = \inf\{ t \in \mathbb{R}^+ \,|\, c(t) = 100 \}$ and $\inf$ denotes the infimum, or the greatest lower bound. The objective is to find the optimal level of distributiveness ($n$):
$\underset{n}{\text{argmin}}\;\;t_f\;.$

\section{Formal Analysis}




We derive the initial coverage rates for the four velocity profiles, ignoring overlaps with other agents.
Let $\mathcal{P}_{0}^{f}$ and $\mathcal{P}_{0}^{i}$ denote the area that an agent had covered after and prior to a given placement, respectively, and $r$ denotes its radius.
An agent covers new area at a rate proportional to its diameter and velocity (see ~\cref{fig:init-coverage}). Therefore, as \small $r = \sqrt{\frac{A}{n\pi}}$, the initial collective coverage rate $\frac{d\mathcal{P}_{0}}{dt}$ of all agents is:
\begin{align*}
            d\mathcal{P}_{0} &\propto n \times  (\mathcal{P}_{0}^{f} - \mathcal{P}_{0}^{i}) \\
            d\mathcal{P}_{0} &\propto n \times  (Vdt 2r + \pi  r^2 - \pi  r^2) \\
            \frac{d\mathcal{P}_{0}}{dt} &\propto n \times 2r \times  v(n)   \propto 2 \times \sqrt{\frac{An}{\pi}} \times v(n) 
\end{align*}
\begin{table}[t]
\centering
\renewcommand*{\arraystretch}{1.2}
\caption{Initial coverage rates for various velocity profiles (from \cref{fig:vel_prof}).} 
\label{tab:coverage_analysis} 
\begin{tabular}{lp{0.3cm}lp{0.3cm}l}
\toprule
\textbf{Constant} & & \textbf{Linear} \\
\midrule
$ \frac{d\mathcal{P}_{0}}{dt} \propto 2\sqrt{\frac{An}{\pi}} \cdot V_0 \propto \sqrt{n} $ & &
$ \frac{d\mathcal{P}_{0}}{dt} \propto 2\sqrt{\frac{An}{\pi}} \cdot \Big(V_0 - \gamma(n-1)\Big) \propto a\sqrt{n} - bn^{3/2} $ \\
\bottomrule
\toprule
\textbf{Radius} & & \textbf{Area} \\
\midrule
$ \frac{d\mathcal{P}_{0}}{dt} \propto 2\sqrt{\frac{An}{\pi}} \cdot \frac{V_0}{\sqrt{n}} \propto \mathcal{P}_c $ & &
$ \frac{d\mathcal{P}_{0}}{dt} \propto 2\sqrt{\frac{An}{\pi}}  \cdot\frac{V_0}{n} \propto \frac{1}{\sqrt{n}}$ \\
\bottomrule
\end{tabular}
\end{table}
Table \ref{tab:coverage_analysis} presents the initial coverage rates for all velocity profiles.


\section{Simulation Results}
This section uses computer simulations to evaluate how different levels of distributiveness affect coverage performance for disk-shaped agents (see \cref{sec:problem}) under no gradual motion, varying velocity profiles, failure rates, and collisions. Each configuration was tested across 30 different initial conditions.  The cumulative coverage percentage~$c(t)$ achieved by the agents was recorded using tools from the OpenCV library~\cite{opencv_library}. We use a no-split agent velocity: $V_0 = 0.005$~units/step, a linear velocity-profile slope: $\gamma = 4 \times 10^{-6}$, and an environment of size: $p \times p = 1 \times 1$~sq.~units discretized into $m \times m = 1000 \times 1000$ cells. The total footprint is fixed at $A = \pi \times 0.1^2$~sq.~units.

\subsection{Effect of No Gradual Motion}
\begin{figure}
  \centering
  \subfloat[]{
        \includegraphics[width=0.35\textwidth]{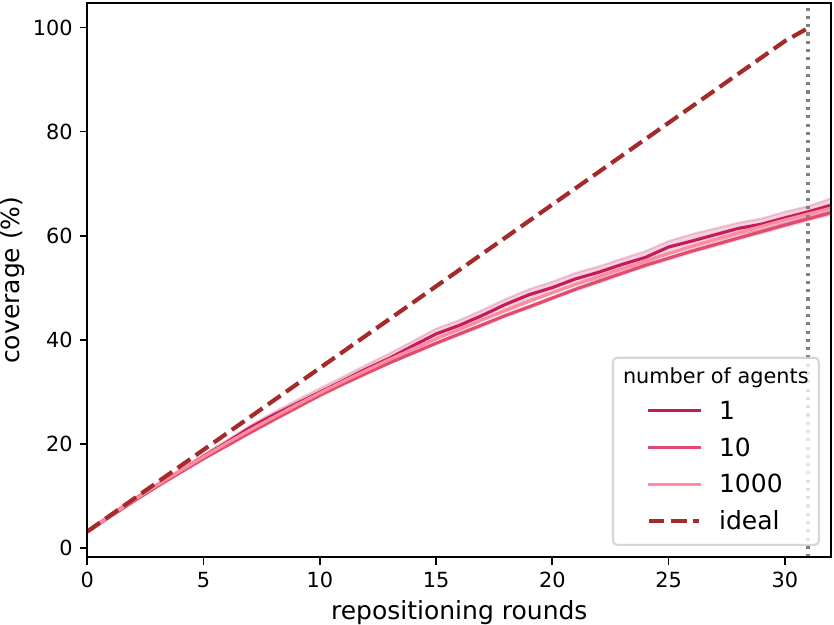}
        \label{fig:no_motion}
  }
  \subfloat[]{
        \includegraphics[width=0.27\textwidth]{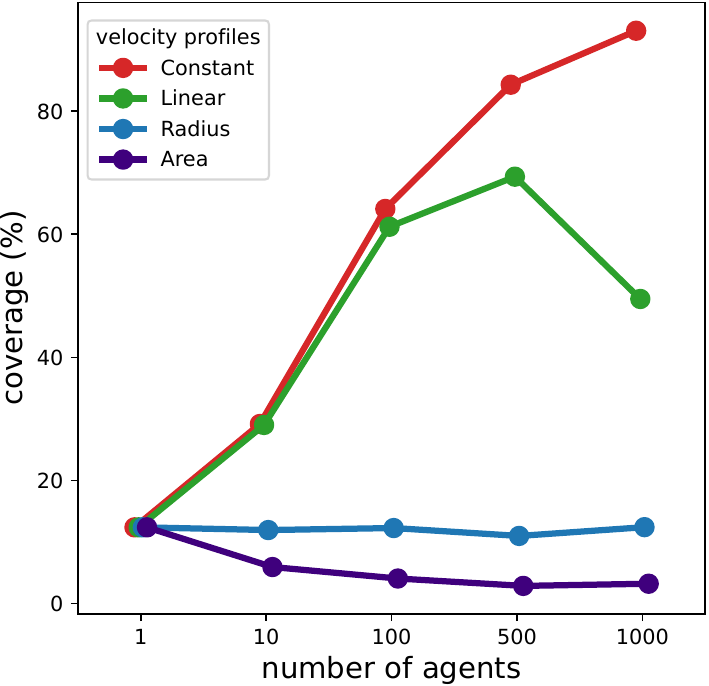}
        \label{fig:coverage_init}
  }
  \caption{ (a) Coverage trends when agents instantly relocate to random positions (b) Coverage at 100th simulation step with various velocity profiles.}
\end{figure}
We investigate the scenario in which agents can instantly appear in any random position in the environment without undergoing gradual movement. This experiment isolates the effect of resource splitting from other influences, such as velocity profiles and physical collisions altering the course. These experiments use the parameters described above, except the total number of cells is increased to $10^4 \times 10^4 $. Moreover, coverage is reported once per \emph{repositioning round}, that is, each time the entire group has been repositioned. As seen in \cref{fig:no_motion}, the coverage trends are identical across all group sizes. Coverage initially even match ideal rates, corresponding to perfect avoidance of revisiting explored regions (i.e., coverage rate of $A$ per round).
However, after a few rounds, the coverage achieved for every group size begins to deviate from the ideal rates, as overlaps with already visited regions increase and the probability of visiting unvisited cells decreases proportionally to the overall coverage.

\subsection{Effect of Velocity Profiles}
We investigate the effects of velocity profiles (see \cref{fig:vel_prof}). We first examine the initial performance, reporting coverage percentages at the 100th simulation step. As shown in \cref{fig:coverage_init}, the coverage trends match those of the formal analysis in \cref{tab:coverage_analysis}.
In particular:

\textbf{Constant:} Coverage increases monotonically with the number of splits. For the group sizes considered, $1000$ agents achieve the highest coverage, indicating that the optimal group size tends toward $\infty$ agents, consistent with \cref{tab:coverage_analysis}. In practice, this suggests distributing resources among as many agents as possible.

\textbf{Linear:} 
Of the group sizes tested, $n=500$ agents achieve the highest coverage, implying that the optimal group size lies between the two extremes ($1$ and $\infty$). To understand this behavior, consider the relationship between coverage rate and $\sqrt{n}$ from \cref{tab:coverage_analysis}, which takes the form $y = ax - bx^{3}$, where $y$ is the coverage rate, $x = \sqrt{n}$, $a,b>0$ and $n\geq1$. This cubic equation attains its local maximum at $x = \sqrt{\frac{a}{3b}}$, or equivalently, at $n = \frac{a}{3b}$. For the parameters in our setup, this corresponds to $n = 417$ (and $n=500$ is the closest from the set of tested agent numbers).

\textbf{Radius:} Coverage remains nearly constant across all group sizes. This indicates that an increase in benefit from distributiveness compensates the reduced mobility of an individual agent at larger group sizes.

\textbf{Area:} Coverage decreases as the group size ($n$) increases, showing that it is optimal to keep all resources within a single agent.
\begin{figure}[t]
    \centering
\includegraphics[width=0.7\linewidth]{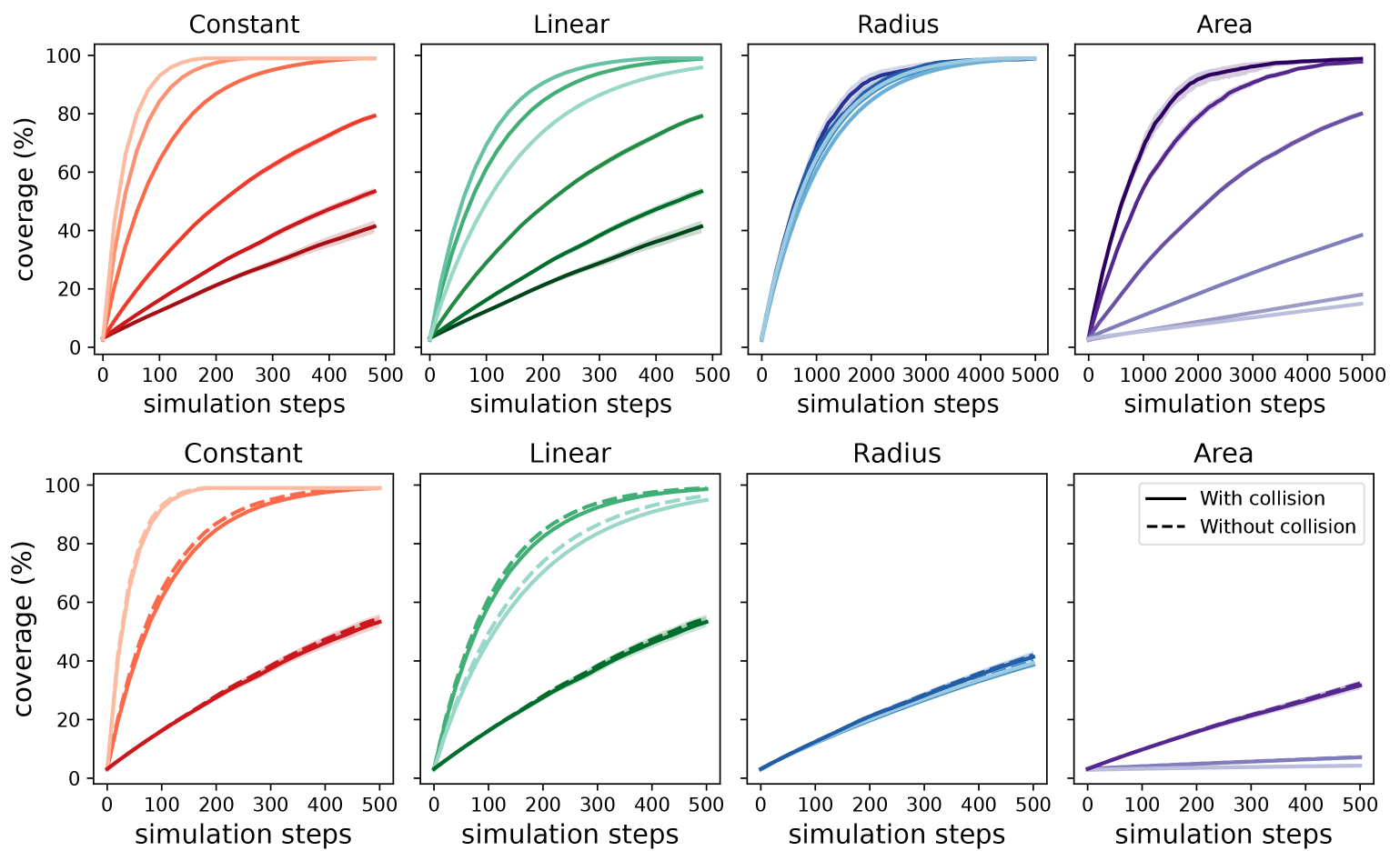}    
    
    \caption{Coverage trends across velocity profiles. The top row shows collision-free results for 1, 2, 10, 100, 500, and 1000 agents; the bottom row shows results with collisions for 2, 100, and 1000 agents. In each profile, colour intensity encodes group sizes, with darker shades indicating fewer agents.}
    \label{fig:coverage_collision}
\end{figure}

These trends persist beyond the initial steps, as seen in the top row of \cref{fig:coverage_collision}. In the constant profile, coverage increases from the darkest to the lightest shade of red (group sizes $1 \rightarrow 1000$). For the linear profile, the performance order is as follows: the second-lightest green ($500$ agents) achieve the highest coverage, followed by $100$, $1000$, $10$, $2$, and $1$ agents that perform in decreasing order thereafter. In the radius profile, all shades remain nearly identical, reflecting the uniform coverage across group sizes. In the area profile, coverage decreases from the darkest to the lightest purple ($1 \rightarrow 1000$).


\subsection{Effect of Collisions}
While the previous sections discarded the effect of collisions, this section quantifies their impact. In multi-robot systems, when robots are about to collide, they typically slow down and rotate away from each other until the situation is resolved, after which they resume their task at the original velocity~\cite{goldberg1997interference}. 
These robot-robot interferences commonly accumulate to long-term~\cite{schroeder2019balancing} and global effects~\cite{schroeder2019balancing,HamannScalability22}. 
We approximate this behavior by decreasing an agent's velocity proportionally to the overlap between its footprint and that of any colliding neighboring agents, while ignoring heading corrections. To prevent deadlocks, a residual velocity is maintained even in the case of full overlap.


We evaluate the effect of collisions on all velocity profiles. As shown in the bottom row of \cref{fig:coverage_collision}, there is no significant decrease in coverage for experiments where the total footprint split over $n$ agents is $A = \pi \times 0.1^2$~sq.\ units, and the coverage trends closely resemble those observed in the top row of \cref{fig:coverage_collision}. For larger footprints, deviations may increase due to the higher impact of an individual overlap with other agents. 

Notably, unlike classical scalability studies~\cite{HeikoSuper18,SomaCongestion23}, where a fixed footprint per agent leads to congestion-induced performance degradation as group size grows, agent density remains constant here. Nevertheless, except for the constant velocity profile, increasing $n$ still reduces agent velocities, emulating congestion like effects.

\subsection{Effect of Failures}

Typically, in swarm robotics, increasing the number of robots introduces miniaturization-related risks~\cite{yang2022review,pfeiffer2017fundamental} that may cause agents to fail over the course of the experiment~\cite{caprari2000fascination,chin2023minimalistic}. In this section, we model such effects through a failure rate $k(n)$ that depends only on $n$. At each simulation step, we sample a uniform random  number~$r\sim \mathcal{U}(0,1)$ for every agent, and if $r<k(n)$, the agent ceases motion
and does not contribute to coverage thereafter. We define $k(n) = \beta(1 - \frac{1}{n^{\alpha}})$,
which is a non-linear, monotonically increasing function with a horizontal asymptote. For any $\alpha > 0$ and $\beta \geq 0$, the failure rate for $n=1$ is $0$. Hence, a sole robot would always be assumed fault-free. $\beta>0$, determines the maximum failure rate for $n \to \infty$. $\alpha$ controls how rapidly the failure rate approaches $\beta$ as $n$ increases (except for $n=1$). 

The experiment compares the performance of a single agent with zero failure probability to that of multi-agent groups with failure rate $k(n)$. We use $\beta = 0.1$ and $\alpha \in \{0.01,\, 0.1\}$ in this study. All agents have constant velocity $V_0=0.005$ units/step.

\begin{figure}[t]
    \centering
    \includegraphics[width=0.7\linewidth
    ]{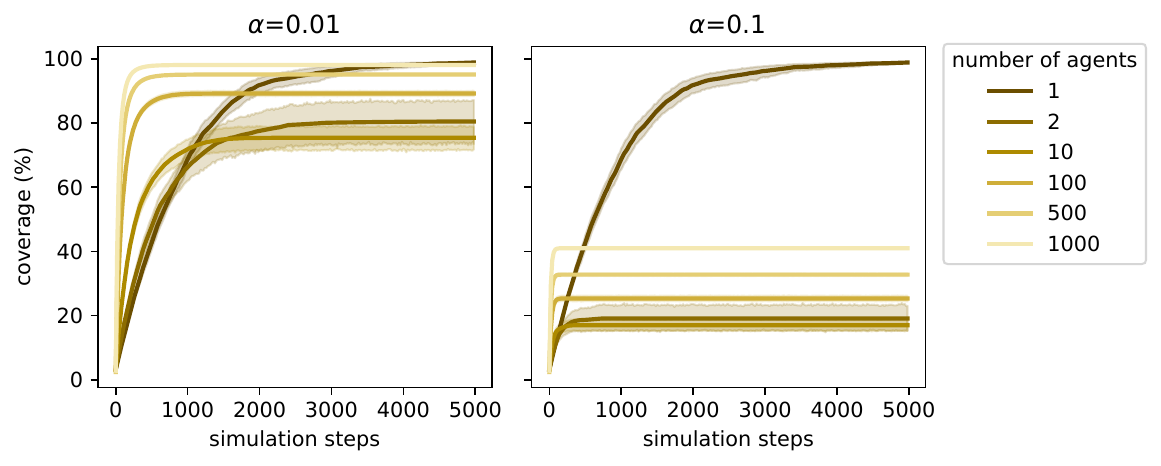}
    \caption{Coverage trends for different failure rates (left: $\alpha =0.01$ and failure rate $k<0.007$; right: $\alpha =0.1$ and failure rate $k\le 0.05$). $\alpha$ controls the rate at which failure rate approaches its maximum failure rate $\beta=0.1$, where $k(n) = \beta(1 - \frac{1}{n^{\alpha}})$.}
    \label{fig:reliablity}
\end{figure}

As shown in \cref{fig:reliablity}, for small $\alpha = 0.01$, large group sizes (e.g., $1000$ agents) remain robust to failures and consistently outperform a single agent, whereas small groups eventually fall behind due to a larger proportion of agents failing. 
For the large $\alpha = 0.1$ and for the same $n$, the failure rate approaches $\beta$ more rapidly, causing agents to fail earlier in the simulation. Under this condition, a single agent overtakes every group size $n$ even before covering a significant proportion of the environment.

\section{Conclusions}
In this work, we formulated the split over $n$ resource sharing problem. It explores a fundamental question in swarm robotics: What level of distributiveness is optimal? We presented a case study built around a simple coverage task. We formally derived the initial coverage rates for a range of velocity profiles.
Through a series of computer simulations, we showed that technological constraints due to miniaturization, such as reduced mobility and increased failure rates, shift the optimal level of distributiveness.
Depending on the type of robots available, our findings could inform system designers in determining whether a single highly capable robot, several moderately capable robots, or a large group of smaller, less capable robots would be most effective for a coverage task. Future work will investigate the performance of heterogeneous robot swarms, combining highly capable and less capable robots.
We will also study how these findings translate to physics-based simulations, assessing whether the observed trends
hold. 


\subsubsection{\ackname}
This research was supported in part by the Mitacs Globalink Research Award, BMBF (Robotics Institute Germany; grant no. 16ME1001) and EU Horizon Europe Framework Programme (``OpenSwarm''; grant no.~101093046). 

\subsubsection{\discintname}
The authors declare no competing interests.

%
%
\bibliographystyle{splncs04}
\bibliography{mybibliography}

\end{document}